# The Dependency Divide: An Interpretable Machine Learning Framework for Profiling Student Digital Satisfaction in the Bangladesh Context

Md Muhtasim Munif Fahim[1], Humyra Ankona[2], Md Monimul Huq[3], Md Rezaul Karim[1, *]

[1]Data Science Research Lab, Department of Statistics, University of Rajshahi, Rajshahi – 6205, Bangladesh
[2]Department of Veterinary & Animal Sciences, University of Rajshahi, Rajshahi – 6205, Bangladesh
[3]Department of Statistics, University of Rajshahi, Rajshahi – 6205, Bangladesh

Email: fahim.stat.ru@gmail.com; humyra.dvm@gmail.com; mhuq75@gmail.com; mrkarim@ru.ac.bd

**\*Corresponding Author**

## Abstract

**Background**: While digital access has expanded rapidly in resource-constrained contexts, satisfaction with digital learning platforms varies significantly among students with seemingly equal connectivity. Traditional digital divide frameworks fail to explain these variations, assuming monotonic relationships between access, engagement, and outcomes.

**Purpose**: This study introduces the "Dependency Divide"—a novel framework proposing that highly engaged students become conditionally vulnerable to infrastructure failures, challenging assumptions that engagement uniformly benefits learners in post-access environments.

**Methods**: We conducted a cross-sectional study of 396 university students in Bangladesh using a three-stage analytical approach: (1) stability-validated K-prototypes clustering to identify student profiles, (2) profile-specific Random Forest models with SHAP and ALE analysis to determine satisfaction drivers, and (3) formal interaction analysis with propensity score matching to test the Dependency Divide hypothesis. Design-weighted analyses ensured population-level generalizability.

**Results**: Three distinct profiles emerged: Casually Engaged (58%), Efficient Learners (35%), and Hyper-Engaged (7%). A significant interaction between educational device time and internet reliability ($\beta = 0.033$, $p = 0.028$) confirmed the Dependency Divide: engagement increased satisfaction only when infrastructure remained reliable, but amplified frustration under poor connectivity. Hyper-Engaged students showed greatest vulnerability despite—or because of—their sophisticated digital workflows. Mediation analysis revealed that 15% of engagement's effect on platform satisfaction operated through heightened awareness of infrastructure failures. Policy simulations demonstrated that targeted reliability improvements for high-dependency users yielded 2.06 times greater returns than uniform interventions.

**Conclusions**: In fragile infrastructure contexts, capability can become liability. Digital transformation policies must prioritize reliability for dependency-prone users, establish contingency systems, and educate students about dependency risks rather than uniformly promoting engagement.

# Introduction

Digital learning technologies have changed Higher Education; however, the advantages of such technologies continue to be unevenly distributed. For example, in resource-poor regions, this inequality manifests itself in a contradictory manner: increased Internet access does not translate to equal educational results - what we refer to as the "post-access paradox" (Okai-Ugbaje, 2020; Badiuzzaman et al., 2021). Bangladesh is an excellent example of this contradiction. Mobile internet penetration had achieved nearly universal levels among university students during the COVID-19 pandemic; yet, the degree of satisfaction with digital learning platforms varied significantly among students with what appeared to be the same level of access.

The inability of traditional digital divide models to explain the variations in satisfaction among students with apparently equal levels of access is due to the fact that they are based on a paradigm of monotonic relationships. That is, more access, more skills, and more engagement should all equate to greater educational outcomes.

We argue that this paradigm fails in contexts where the reliability of infrastructure is a concern. There is evidence across a variety of resource-poor contexts that it is not just the existence of access to the Internet, but the reliability of the connection that determines the outcome. Malaysian students indicate that the main barrier to learning is connectivity interruptions (Wong et al., 2025). Similarly, Indonesian medical students indicate that unreliable internet is their greatest concern regarding pedagogy (Ortiz Riofrio et al., 2024). A similar finding is reported by Salas-Pilco et al. (2022): Latin American students identified unstable connections, not device ownership, as the largest barrier to engagement with course materials during the COVID-19 pandemic. Of particular relevance to our argument, Ho et al. (2021) demonstrated that despite possessing technical ability and access to devices, Hong Kong students' overall satisfaction with emergency remote learning was neutral (M = 4.11 on a seven-point scale), suggesting that there were unmeasured quality-of-infrastructure factors that mediated satisfaction. Furthermore, Freyens and Gong (2022) demonstrated the consequences of this - Australian students who relied heavily on lecture recordings performed worse than those who attended live classes, indicating dependence risks when connectivity issues prevent access to course content.

Therefore, these findings suggest a disturbing possibility - that the most digitally engaged students will be the ones who become the most vulnerable to infrastructure failures, not the least. We propose the concept of the Dependency Divide as a new form of inequality that exists in the context of post-access environments. While traditional divides focus on who lacks access or skills, the Dependency Divide identifies conditional vulnerability to infrastructure failure among students who are the most digitally engaged. Those students who create complex workflows to facilitate their learning through reliable connectivity (e.g. attend synchronous sessions, view recorded lectures, participate in online discussions) will experience greater disruption than others when infrastructure fails. In this way, capability can become a liability. The relationship between engagement and satisfaction is not linear but conditional: positive when infrastructure remains reliable, potentially negative when reliability fails.

To test this hypothesis, research must move away from the descriptive profiling currently prevalent. Student heterogeneity is recognized in research (Brown et al., 2022; Gašević et al., 2015), but profiling studies are typically descriptive and do not connect these differences to potential policy interventions. Institutional leaders require profile-specific understandings of the factors that drive satisfaction for each group and how these factors interact with the quality of infrastructure. Additionally, few studies employing interpretable machine learning methods exist that can explain why predictions are made and provide insight into which interventions would be effective (Joksimovíc et al., 2017; Ifenthaler & Yau, 2020). Even fewer studies have employed both unsupervised clustering and supervised prediction while preventing data leakage - a critical methodological gap when developing profile-specific models.

We have addressed the above-mentioned shortcomings through an empirical examination of 396 university students in Bangladesh, where there is a growing number of Internet-connected users, yet a large proportion of them still face many barriers related to the availability of Internet infrastructure. The researchers have asked three questions: (RQ1) What are the different types of student profiles based on how they have used and accessed the Internet? (RQ2) What are the factors that most strongly determine overall satisfaction and what are those factors within each type of student profile? (RQ3) Based on their profile, what are the specific policies that institutions can put in place to support students in developing countries who do not have sufficient resources?

We have developed a three-stage analytical model to address these questions; (Stage 1) Stability-validated K-prototypes clustering, which is suitable for analyzing educational data that has multiple formats (quantitative and qualitative); (Stage 2) Profile-specific Random Forest models with SHAP and ALE analysis to analyze the relationships between the different types of student profiles identified and their levels of satisfaction with the online educational experience; and (Stage 3) Formal interaction analysis to test if the different types of student profiles are moderated by differences in Internet infrastructure.

The results of this study identified three types of profiles: Casually Engaged (58%), Efficient Learner (35%), and Hyper-Engaged (7%). The researcher also demonstrated that the three types of profiles were different not only in terms of the amount of time they spent using and accessing the Internet, but also in terms of how vulnerable they were to Internet outages. The researcher provided the first empirical evidence for the Dependency Divide and demonstrated that engagement creates a risk for highly engaged students in unstable Internet environments rather than providing a benefit. This challenge's traditional assumptions in technology acceptance studies and demonstrate that engagement is not uniformly beneficial.

Finally, we translated their findings into actionable recommendations for educators and policymakers working in low-resource settings; infrastructure investment should be prioritized for students who rely heavily on the Internet; contingency plans and backup systems should be established so that students can continue to learn while the Internet connection is unavailable due to an outage; and students need to be made aware of the risks associated with relying on the Internet too much.

# Methods

## Study context and design

We conducted this study at a large public University in Bangladesh; it is a Global South setting with relatively widespread nominal connectivity that is experiencing continuing infrastructural issues. A cross-section survey collected data from students' demographic characteristics, their digital accessibility, their use of devices, as well as their level of satisfaction for items on an institutionally supported platform and infrastructure. This analysis employed a three-phase analytical strategy — psychometric validation, unsupervised segmentation, and supervised modeling including causal extensions — to provide explanatory insights into student's digital experiences and generate profile specific policy recommendations, while giving up some predictive performance. All of these analyses were run in a reproducible environment using Python 3.12 (scikit-learn 1.6.1, XGBoost 3.0.5, kmodes 0.12.2); a single global random seed (42) was used to ensure consistent results from all stochastic procedures.

## Sampling, survey design, and weighting

The target population comprised undergraduate and graduate students enrolled at the university during the data collection period. A stratified two-stage cluster design was used, with 31 class sections as primary sampling units and strata defined by academic year and program. Base weights were defined as the inverse of selection probabilities, adjusted for non-response within strata, and post-stratified to known gender and year distributions from institutional records.

The final survey yielded 447 consented students, of whom 396 provided complete responses on the core variables and form the full analytic sample for psychometrics, clustering, and descriptive statistics. For supervised modeling, a listwise-complete subsample of 340 students was created by excluding cases with missing outcome data, with attrition driven primarily by missing PlatformHelpfulness responses. A further reduced sample of 315 students was used for interaction models requiring complete data on TimeOnDeviceEducation, InternetReliability, and covariates. All population-level estimates and model metrics are design-weighted, with section-cluster–robust standard errors used for inference wherever applicable.

## Measures

Demographic predictors included Age (years) and Gender. Digital access variables captured computer ownership (HasComputer), self-reported InternetReliability on a five-point scale (Very bad to Very good), and proxies for socio-economic status such as monthly FamilyExpenditure and individual MonthlyExpenditure categories. Usage measures included total TimeOnDevice (hours/day) and a cleaned educational device time variable (TimeOnDeviceEducation_Cleaned), standardized to hours per day from heterogeneous free-text entries using regular-expression–based parsing rules.

Satisfaction outcomes comprised three ordinal variables: InternetReliability (perceived reliability), PlatformHelpfulness (perceived usefulness of institutional platforms), and TechForLearning (frequency of using technology for learning), each measured on a five-level ordered scale. An additional TechCollaboration item was initially included in the satisfaction battery but removed after psychometric analysis indicated it did not cohere with the other items, as described below.

## Data preprocessing

A unified preprocessing pipeline was implemented using scikit-learn's `ColumnTransformer` and persisted for reproducibility. Numerical predictors underwent median imputation followed by standardization, with winsorization of TimeOnDevice and TimeOnDeviceEducation_Cleaned at the 1st and 99th percentiles to reduce undue influence of extreme values while preserving genuine high-use behavior. Ordinal predictors were imputed using the most frequent category and encoded with ordered integer codes reflecting substantive category order. Nominal predictors were imputed using the most frequent category and encoded via one-hot indicators, with rare levels (less than 2% frequency) pooled into an "Other" category to stabilize estimates.

All preprocessing steps were fit separately within training folds during cross-validation, and the resulting transformers were then applied to held-out test folds to prevent information leakage. Satisfaction items were always treated as outcomes and never included among predictors in any supervised model to avoid label leakage into feature construction.

## Analytical framework

### Psychometric analysis

Because the satisfaction items are ordinal Likert-type responses, internal consistency was assessed using Cronbach's alpha computed on a Spearman rank correlation matrix, complemented by ordinal (polychoric) reliability estimates. The four-item scale (InternetReliability, TechForLearning, PlatformHelpfulness, TechCollaboration) exhibited very low reliability (Spearman alpha $\approx 0.228$), and alpha-if-item-deleted analyses together with corrected item–total correlations flagged TechCollaboration as poorly aligned with the other items. Subsequent analyses therefore modeled InternetReliability, PlatformHelpfulness, and TechForLearning as separate outcomes rather than as a single composite satisfaction index.

### Unsupervised segmentation

To answer RQ1 on student profiles, k-prototypes clustering was applied to mixed-type predictors (demographics, access, and usage variables) using the Huang distance, which combines Euclidean distances for numeric variables and simple mismatches for categorical variables. Candidate solutions with $k \in \{2, \ldots, 8\}$ were compared using inertia, a weighted silhouette index, and the Calinski–Harabasz index, with $k = 3$ selected based on the highest silhouette ($\approx 0.142$) and favorable CH values alongside interpretability of the resulting clusters. Negative silhouette scores for some k were expected due to overlap in mixed-type data and were interpreted accordingly.

Clustering hyperparameters were set to `n_init = 20`, `max_iter = 300`, and `init = "Huang"`, with the numeric–categorical trade-off parameter $\gamma$ determined via the Huang heuristic to balance numeric variance and categorical mismatches.

Design weights could not be incorporated directly into the clustering algorithm, but were applied post hoc when estimating cluster prevalence and confidence intervals.

Cluster stability was evaluated via bootstrap consensus clustering with 1000 resamples, each including 80% of cases drawn without replacement. For each cluster, a median Jaccard coefficient and interquartile range quantified the proportion of cases consistently reassigned to the same cluster across bootstrap samples. Clusters corresponding to Efficient Learners and Casually Engaged students showed good stability (median Jaccard $\approx 0.58$), whereas the Hyper-Engaged segment exhibited lower stability (median Jaccard $\approx 0.16$), supporting its interpretation as a heterogeneous high-use fringe. Cluster labels and descriptive profiles were derived from design-weighted means and proportions for key variables summarized in the cluster profile tables.

## Supervised modeling and interpretability

To address RQ2 on drivers of satisfaction, separate supervised models were trained for each outcome (InternetReliability, PlatformHelpfulness, TechForLearning) using Random Forest, Gradient Boosting, and XGBoost, with a proportional-odds ordinal regression serving as a baseline. Models were estimated on the 340-case modeling subsample, using StratifiedGroupKFold cross-validation with section as the grouping variable to ensure that students from the same class section never appeared in both training and test folds. The number of folds (between 2 and 6) was chosen per outcome to maintain representation of all outcome categories and class years in each fold.

Design weights were passed as sample_weight to Random Forest and XGBoost models during training and to performance metrics for all algorithms, while the ordinal regression baseline included a random section intercept to partially account for clustering but could not incorporate weights directly into estimation. Performance was evaluated primarily using design-weighted Quadratic Weighted Kappa (QWK), with Macro-F1 as a secondary metric to capture class-wise balance. Final metrics reported in the Results are outer-fold means and standard deviations for each algorithm–outcome combination.

Global model interpretability relied on SHAP values and Accumulated Local Effects (ALE) plots computed on outer-fold test sets. For each outcome, SHAP values were aggregated from one-hot encodings back to parent features and then rank-aggregated across folds using Borda aggregation to obtain stable feature importance rankings. ALE plots were used to visualize the marginal influence of key drivers such as TimeOnDeviceEducation_Cleaned, TimeOnDevice, and Age while accounting for feature correlations, with particular attention to non-linear dose–response patterns. These interpretability tools informed the selection of variables for interaction and causal analyses rather than serving as standalone inferential evidence.

## Statistical test of the Dependency Divide

To formally test the Dependency Divide hypothesis, a design-weighted weighted least squares (WLS) model predicted TechForLearning (treated as a quasi-continuous 1–5 score) from TimeOnDeviceEducation_Cleaned, InternetReliability, and their interaction. The model can be written as

$$\begin{aligned}\text{TechForLearning}_i &= \beta_0 + \beta_1 \text{TimeEdu}_i \\&+ \beta_2 \text{InternetRel}_i \\&+ \beta_3 (\text{TimeEdu}_i \\&\times \text{InternetRel}_i) + \epsilon_i,\end{aligned}$$

with survey design weights applied and robust standard errors clustered at the section level. The primary parameter of interest was the interaction coefficient $\beta_3$, tested at $\alpha = 0.05$, with 95% confidence intervals used to quantify uncertainty. Predicted values from this model were later used to generate marginal effects plots stratified by high versus low InternetReliability and by cluster to visualize the Dependency Divide across student profiles.

## Propensity score matching for causal robustness

Because high educational device use may be correlated with demographic and socio-economic factors that also influence satisfaction, propensity score matching (PSM) was used to probe the causal robustness of the interaction between TimeOnDeviceEducation_Cleaned and InternetReliability. For these analyses, high versus low TimeOnDeviceEducation_Cleaned was treated as a binary exposure, and propensity scores were estimated via logistic regression including Age, Gender, SES (FamilyExpenditure), ParentEducation, and Class. Students were matched using 1:1 nearest-neighbor matching with a caliper of 0.024, corresponding to 0.25 times the standard deviation of the propensity scores, yielding a matched sample of 249 students and a match rate of 99.4%.

Covariate balance before and after matching was assessed using standardized mean differences and statistical tests, with results reported in Supplementary Table S1 showing all post-matching p-values exceeding 0.05. Within the matched sample, weighted regression models re-estimated the TimeOnDeviceEducation_Cleaned × InternetReliability interaction for both PlatformHelpfulness and TechForLearning, using the same functional form as the main WLS models but without survey design weights. Comparisons of interaction coefficients and confidence intervals between the original and matched samples were summarized in a causal robustness table to evaluate whether the Dependency Divide effect persisted after adjusting for observed selection bias.

## Mediation analysis

To investigate mechanisms underpinning the Dependency Divide, a mediation model examined InternetReliability as a mediator of the relationship between TimeOnDeviceEducation_Cleaned and PlatformHelpfulness. Two linear regression equations were estimated: one for the mediator (InternetReliability) predicted by TimeOnDeviceEducation_Cleaned and covariates, and one for PlatformHelpfulness predicted by both TimeOnDeviceEducation_Cleaned and InternetReliability. The indirect effect was calculated as the product of Path A (TimeOnDeviceEducation_Cleaned → InternetReliability) and Path B (InternetReliability → PlatformHelpfulness), with statistical significance assessed using bias-corrected bootstrap confidence intervals based on 1000 resamples. The proportion mediated was computed as the ratio of the indirect to the total effect to quantify how much of the engagement–satisfaction relationship operates through perceived reliability.

### Policy simulation and dose–response analysis

To address RQ3 on policy implications, counterfactual simulations used the best-performing supervised models to estimate the impact of improving InternetReliability on PlatformHelpfulness for each student cluster. For each student, predicted PlatformHelpfulness was computed under the observed reliability and under a scenario in which InternetReliability was increased by one standard deviation (≈0.95 points on the five-point scale) while all other predictors were held constant. Cluster-specific average treatment effects and 95% confidence intervals were derived from these predictions and summarized in a forest-plot–ready table, enabling comparison of the effectiveness of reliability improvements across personas.

In parallel, a dose–response analysis examined the relationship between TimeOnDeviceEducation_Cleaned and predicted PlatformHelpfulness under high versus low InternetReliability, using ALE-derived curves and model-based predictions to identify thresholds beyond which additional engagement yields diminishing returns. These analyses were descriptive rather than inferential but tightly coupled to the supervised models and clustering framework described above.

## Qualitative analysis

The qualitative data from the open-ended questions of 396 participants were used for qualitative analysis, with the purpose of providing context to the quantitative results and giving voice to the previously described personas. The qualitative data were analyzed using Braun & Clark's (2006) reflexive thematic analysis method, with the first level being an open coding process that resulted in the development of a preliminary code book. A second independent coder then coded a random 20% subsample of the qualitative data, with the result being high inter-rater reliability (Cohen's kappa > 0.85). Any discrepancies between the coders were resolved through dialogue and the next step was axial coding where the lower-level codes were grouped together into higher-level themes related to infrastructure instability, pedagogy and platform usability. Finally, the results were interpreted based on the cluster profiles identified in the quantitative analyses. Support for the trustworthiness of the qualitative analyses were provided by intercoder reliability assessments, the audit trail of the analysts' decision-making processes and the use of relevant quotes as evidence in the Results.

## Ethics and reproducibility

The study received ethical approval from the relevant institutional review board (to be inserted: IRB/IEC name and protocol ID), and all participants provided informed consent prior to survey completion. Data were anonymized, stored securely, and analyzed in aggregate to minimize risks of re-identification, particularly for small clusters such as the Hyper-Engaged group. All analysis scripts, preprocessing pipelines, and configuration files were maintained under version control, and a container manifest documenting package versions ensures computational reproducibility; de-identified data and code will be made available via an open repository upon publication (insert DOI/link).

# Results

## Measurement and analytic samples

Psychometric analysis of the four satisfaction items yielded a Cronbach's alpha of $\alpha = 0.228$, indicating that the scale does not behave as a single latent construct and justifying item-level modeling rather than aggregation into a composite index. Corrected item–total correlations and alpha-if-item-deleted diagnostics showed that the TechCollaboration item contributed most to the unreliability of the scale, and it was therefore excluded from subsequent analyses of digital satisfaction. The focal outcomes in all following models are the three remaining ordinal variables: InternetReliability, PlatformHelpfulness, and TechForLearning, each treated as a five-level ordered response.

The full weighted analytic sample comprised $n = 396$ students, with design weights reflecting the two-stage cluster sampling and post-stratification to known gender and year distributions at the university. The unsupervised segmentation and descriptive profiling used this full sample, whereas supervised modeling relied on a listwise-complete subsample of $n = 340$ for the three satisfaction outcomes and $n = 315$ for models including all predictors required for interaction analyses. All point estimates reported below are design-weighted, and uncertainty is quantified using section-cluster–robust variance estimators unless otherwise specified.

## Student profiles from unsupervised segmentation (RQ1)

To address RQ1, three student profiles were derived using k-prototypes clustering on mixed-type predictors covering demographics, digital access, and device usage patterns. Candidate solutions with $k \in \{2, \ldots, 10\}$ were compared using elbow, silhouette, and Calinski–Harabasz indices, with the best silhouette score observed at $k = 3$ (0.142) and a favorable separation profile in the Calinski–Harabasz index, supporting a three-cluster solution on both statistical and interpretive grounds.

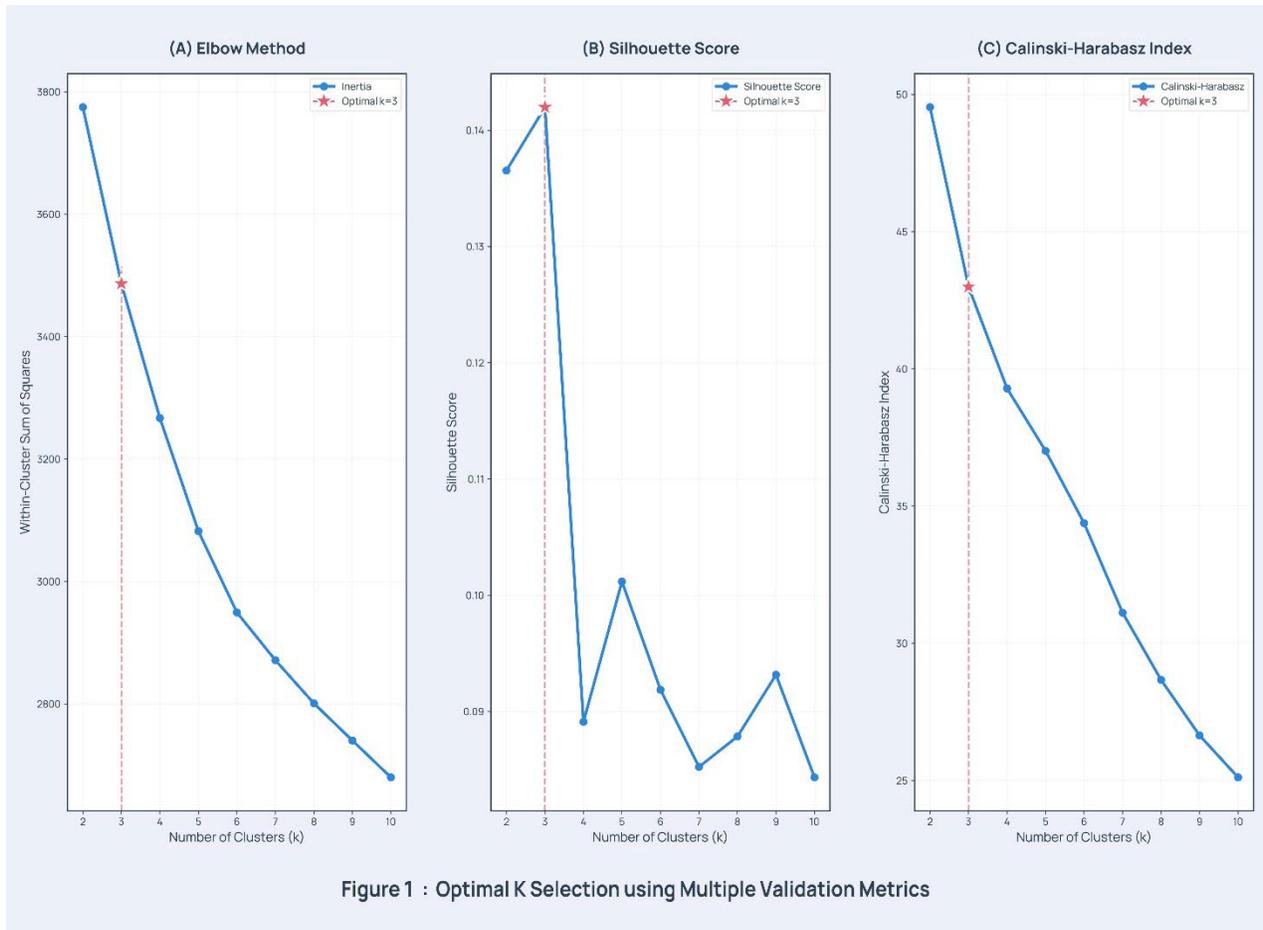

Figure 1 : Optimal K Selection using Multiple Validation Metrics

[Figure 1 – Internal validation of the three-cluster solution (elbow, silhouette, Calinski–Harabasz).]

Bootstrap stability analysis (B = 1000, 80% row subsamples) demonstrated acceptable robustness for two clusters and lower stability for the third. Median Jaccard indices were approximately 0.58 for two clusters and around 0.15 for the remaining cluster, suggesting two well-defined cores and one more heterogeneous high-use fringe. These stability patterns guided the interpretation of clusters as personas rather than rigid, mutually exclusive types, especially for the smallest segment.

Design-weighted prevalences suggested that three profiles do a strong job of capturing the overwhelming majority of the population: a large Casually Engaged segment (≈58.2%, 95% CI [52.9%, 63.4%]), a large Efficient Learner segment (≈34.7%, 95% CI [29.8%, 39.9%]), and a small Hyper-Engaged segment (≈7.1%, 95% CI [4.8%, 10.3%]). Demographic differences in different segments reported in Table 1 (enriched cluster profiles) hint at more variances -especially that Efficient Learners are younger on average and more gender balanced- alongside access and usage differences, including striking differences in total and educational device time. Taken on aggregate total, these differences read as part of the behavioral continuum, with the Hyper-Engaged group representing an extreme high-dependency tail, rather than a cohesive behavioral category.

**Table 1.** Design-weighted profiles of the three student clusters, with omnibus tests and effect sizes.

| Characteristic | C0: Hyper-Engaged | C1: Efficient Learners | C2: Casually Engaged | Test statistic | p-value | Effect size |
| --- | --- | --- | --- | --- | --- | --- |
| Age (years) | 23.25 ± 1.80 | 21.06 ± 1.02*** | 23.83 ± 1.16*** | $F(2,337) = 208.91$ | <0.0001 | $\eta^2 = 0.554$ |
| TimeOnDevice (hrs/day) | 11.83 ± 3.71 | 5.10 ± 2.45*** | 6.20 ± 2.85*** | $F(2,337) = 58.18$ | <0.0001 | $\eta^2 = 0.257$ |
| TimeOnDeviceEducation (hrs/day) | 7.71 ± 2.79 | 2.01 ± 1.15*** | 1.91 ± 1.28*** | $F(2,337) = 186.13$ | <0.0001 | $\eta^2 = 0.544$ |
| Gender (% Female) | 54.2% F | 37.3% F*** | 23.2% F*** | $\chi^2(2) = 13.89$ | 0.001 | Cramér's V = 0.202 |
| InternetReliability (% Bad/Very Bad) | 8.30% | 9.3% * | 9.6% * | $\chi^2(8) = 16.47$ | 0.0361 | Cramér's V = 0.156 |

Notes: *** $p < 0.001$ vs C0; ** $p < 0.01$; * $p < 0.05$ (pairwise comparisons as specified in the text).

## Describing the three student personas

The Casually Engaged profile comprises the majority of students, who report ready access to devices but devote relatively little time to educational activities (around 1.8 hours/day) and show moderate satisfaction across the three outcomes. Quantitatively, this group exhibits moderate total screen time but low educational share, and qualitatively their comments emphasize ambivalence about digital tools, oscillating between recognition of their usefulness and frustration with distraction and pedagogical quality. This pattern suggests that the main constraint for these students is not access per se but the availability of compelling, well-structured learning experiences that justify greater educational investment online.

Efficient Learners have similar total time with devices as Casually Engaged, but a somewhat larger share of that time is educational use ($\approx$1.9 Hrs/day), and they report somewhat greater platform and learning satisfaction. This persona is younger on average and more evenly split by gender, and students self-describe as seeing technology as a useful clarifying tool for difficult course material, accessed through specific search and curated resources. This persona reflects students who are engaging the digital infrastructure in a somewhat focused efficiency-natured way who are likely to be able to take benefit from good platforms without being intensive users overall.

The final persona is Hyper-Engaged. This is a small group but crucial to the analysis: their reports indicate extremely high device use ($\approx$ 11-12 Hrs/day) and very high educational device use ($\approx$ 7Hrs/day), but the satisfaction with platforms and infrastructure is muted or less frequently negative. Diagnostics indicate that this group is heterogeneous (median Jaccard $\approx$ 0.15), but qualitative comments repeatedly note the incredible strain of intensive academic use on fragile infrastructure, describing both hyper-efficiency and the psychological burden of breakages. This constellation of extreme engagement in fragile conditions is the empirical footprint for the Dependency Divide concept presented later in the section.

**Table 2.** Bootstrap Jaccard stability indices for each cluster, with cluster sizes and interquartile ranges.

| Cluster | N | Median Jaccard | IQR | Q1 | Q3 |
| --- | --- | --- | --- | --- | --- |
| C0: Hyper-Engaged | 24 | 0.158 | 0.178 | 0.122 | 0.300 |
| C1: Efficient Learners | 118 | 0.586 | 0.171 | 0.489 | 0.660 |
| C2: Casually Engaged | 198 | 0.578 | 0.179 | 0.496 | 0.674 |

*Notes: Jaccard indices are based on B = 1000 bootstrap resamples with 80% of cases drawn without replacement per iteration.*

# Predictive models of satisfaction (RQ2)

To address RQ2, the study modeled each satisfaction item using Random Forest, Gradient Boosting, and XGBoost classifiers alongside an ordinal regression baseline, with StratifiedGroupKFold cross-validation and design weights applied to both model fitting and metric calculation where supported. Quadratic Weighted Kappa (QWK) served as the primary performance metric, complemented by Macro-F1 to assess class-wise balance, reflecting the ordinal structure and imbalance of the outcomes.

On the corrected modeling pipeline, Random Forest achieved the strongest overall performance for InternetReliability (QWK $\approx$ 0.196, SD $\approx$ 0.127; Macro-F1 $\approx$ 0.255), while for PlatformHelpfulness it reached QWK $\approx$ 0.145 with Macro-F1 around 0.321, and XGBoost performed best for TechForLearning (QWK $\approx$ 0.106; Macro-F1 $\approx$ 0.531). These values are modest in absolute terms, but they represent clear gains over the preliminary analyses and are in line with the difficulty of predicting five-level ordinal satisfaction from a relatively small, heterogeneous sample. As a result, interpretation focuses on the stability and direction of key predictors rather than the pursuit of high predictive accuracy.

**Table 3.** Design-weighted predictive performance (mean ± SD across folds) for each outcome and algorithm; QWK = Quadratic Weighted Kappa.

| Outcome | Model | Mean QWK ± SD | Mean Macro-F1 ± SD |
| --- | --- | --- | --- |
| InternetReliability | Random Forest | 0.196 (±0.127) | 0.255 (±0.050) |
| InternetReliability | XGBoost | 0.171 (±0.128) | 0.271 (±0.089) |
| InternetReliability | Gradient Boosting | 0.101 (±0.065) | 0.243 (±0.026) |
| PlatformHelpfulness | Random Forest | 0.145 (±0.109) | 0.321 (±0.068) |
| PlatformHelpfulness | XGBoost | 0.137 (±0.127) | 0.353 (±0.094) |
| PlatformHelpfulness | Gradient Boosting | 0.095 (±0.284) | 0.335 (±0.113) |
| TechForLearning | XGBoost | 0.106 (±0.142) | 0.531 (±0.090) |
| TechForLearning | Random Forest | 0.022 (±0.151) | 0.481 (±0.093) |
| TechForLearning | Gradient Boosting | −0.064 (±0.061) | 0.396 (±0.033) |

Model-agnostic interpretation based on SHAP values, aggregated across outer-fold test sets and stabilized with Borda aggregation, identified TimeOnDeviceEducation, TimeOnDevice, Age, SES proxies (FamilyExpenditure), and access variables (HasComputer, monthly expenditure) as consistently influential predictors across the three outcomes. For InternetReliability and PlatformHelpfulness, TimeOnDeviceEducation and Age occupied the top ranks, whereas TechForLearning was more strongly associated with TimeOnDeviceEducation, TimeOnDevice, and access variables, reflecting its behavioral nature as a frequency-of-use outcome. table 4 summarizes these stable top-five drivers, which form the backbone for subsequent interaction and causal analyses.

**Table 4.** Most influential predictors per cluster based on mean absolute SHAP values from cluster-wise Random Forest models.

| Cluster | Top feature 1 | Top feature 2 | Top feature 3 | Top feature 4 |
| --- | --- | --- | --- | --- |
| C0: Hyper-Engaged | TimeOnDevice | TimeOnDeviceEducation_Cleaned | Class_3rd year | Gender_Male |
| C1: Efficient Learners | TimeOnDeviceEducation_Cleaned | TimeOnDevice | Class_3rd year | Class_Final year |
| C2: Casually Engaged | TimeOnDeviceEducation_Cleaned | Class_3rd year | Class_Final year | TimeOnDevice |

*Notes: Features are ordered following the largest mean absolute SHAP values within each cluster as reported in the cluster-level importance file.*

Accumulated Local Effects (ALE) plots further clarified the shape of these relationships, showing that educational device time is associated with gains in predicted satisfaction up to around 2 hours/day, after which marginal returns flatten or even decline slightly depending on the outcome. For Age, ALE curves suggested a non-linear pattern with relative disadvantage at the youngest ages and a plateau in the mid-twenties, consistent with a maturation effect in digital study strategies. These non-linearities motivated the explicit testing of interaction terms to probe whether the benefits of educational engagement are contingent on infrastructural conditions.

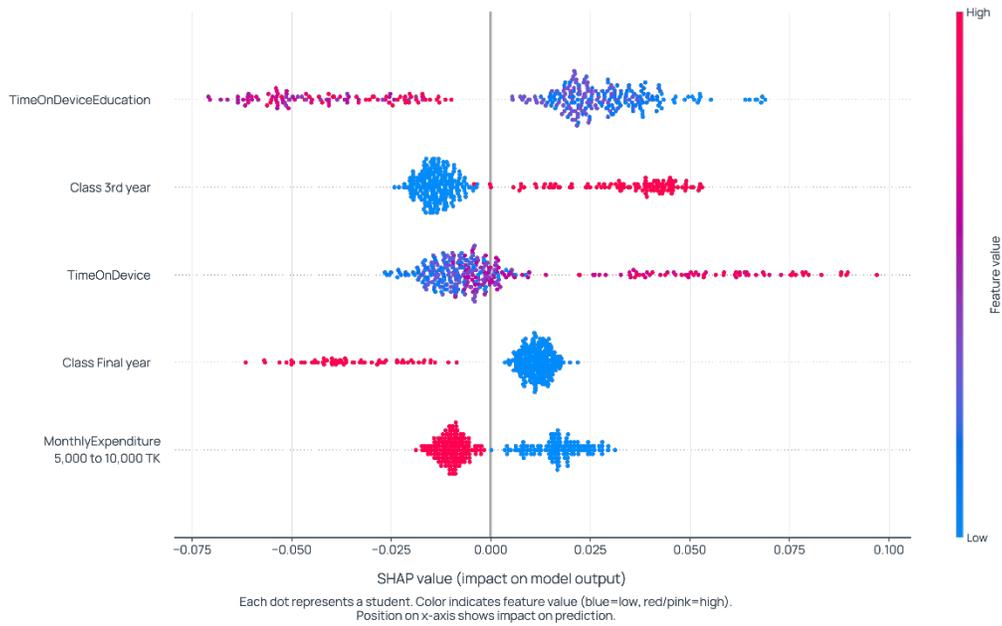

Feature Importance for Predicting "Often" (TechForLearning)

## Accumulated Local Effects (ALE) for Technology Use in Learning

ALE plots show how features affect predictions on average, accounting for feature correlations.

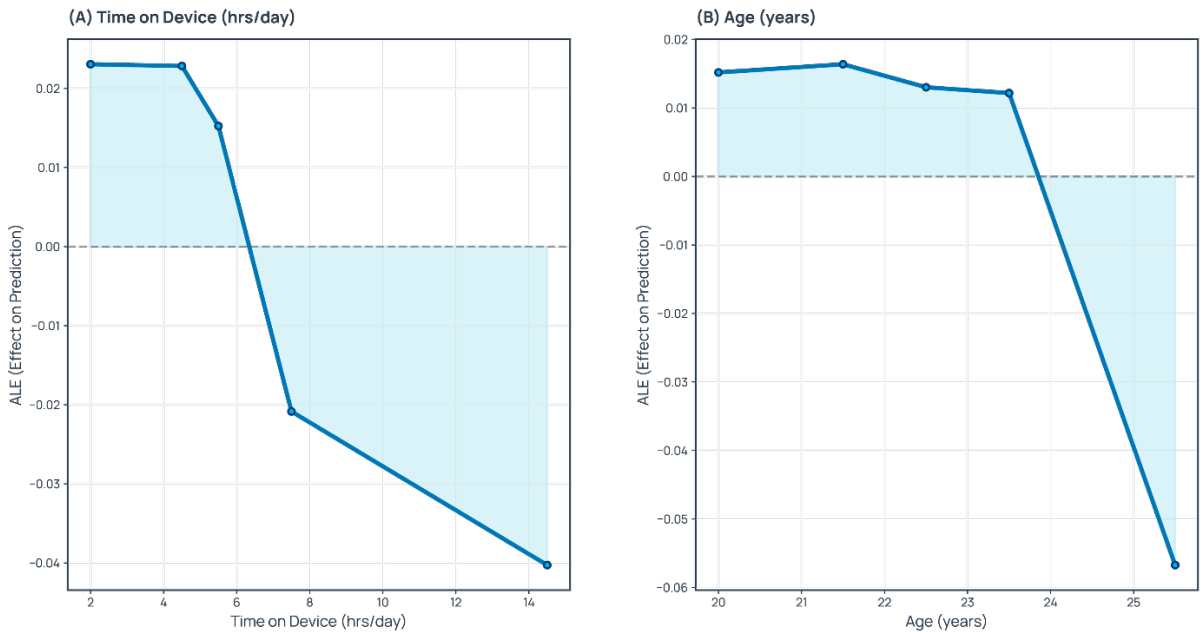

[Figure 2– SHAP summary plot and ALE curves for TimeOnDeviceEducation and Age.]

## Statistical evidence for the Dependency Divide (RQ2)

The Dependency Divide hypothesis posits that the effect of educational device time on satisfaction is conditional on infrastructure reliability, rather than additive, such that the most engaged students are also the most vulnerable when connectivity is fragile. To test this, a design-weighted weighted least squares model was estimated with TechForLearning as the outcome and predictors including TimeOnDeviceEducation, InternetReliability, and their interaction, controlling for clustering via robust standard errors.

Results supported the hypothesized dependency pattern: the interaction term between TimeOnDeviceEducation and InternetReliability was positive and statistically significant ($\beta \approx 0.033$, 95% CI [0.004, 0.062], $p \approx 0.028$), while both main effects were non-significant. The absence of significant main effects indicates that neither higher engagement nor better reliability alone is sufficient to raise satisfaction, and that gains arise instead from their combination, which is precisely the conditional structure implied by the Dependency Divide. In other words, educational engagement pays off only for students who experience sufficiently reliable connectivity, whereas for those with poor reliability, additional device time offers little benefit and may even amplify frustration.

**Table 5.** Design-weighted interaction of educational device time and internet reliability predicting TechForLearning.

| Predictor | β | SE | 95% CI (lower) | 95% CI (upper) | p-value | Interpretation |
| --- | --- | --- | --- | --- | --- | --- |
| TimeOnDeviceEducation (main effect) | −0.0588 | 0.0384 | −0.1345 | 0.0168 | 0.1270 | Direct effect of educational time |
| InternetReliability (main effect) | −0.0247 | 0.0426 | −0.1084 | 0.0591 | 0.5628 | Direct effect of internet quality |
| TimeOnDevice × Internet (interaction) | 0.0331 | 0.0150 | 0.0036 | 0.0625 | 0.0278 | Dependency Divide effect |

Predicted values from this model, plotted across the observed range of TimeOnDeviceEducation for high versus low InternetReliability, reveal a clear divergence: under good or very good internet conditions, satisfaction increases with educational device time up to roughly the 2–2.5 hours/day range, whereas under bad or workable conditions the slope is flat or slightly negative. Cluster-stratified plots further show that this divergence is steepest for the Hyper-Engaged cluster, moderate for Efficient Learners, and smallest for the Casually Engaged group, mirroring their differing levels of dependency on digital tools for study. This pattern provides a quantitative and visual anchor for interpreting the Hyper-Engaged profile as a dependency-vulnerable fringe rather than simply an intensive-use group.

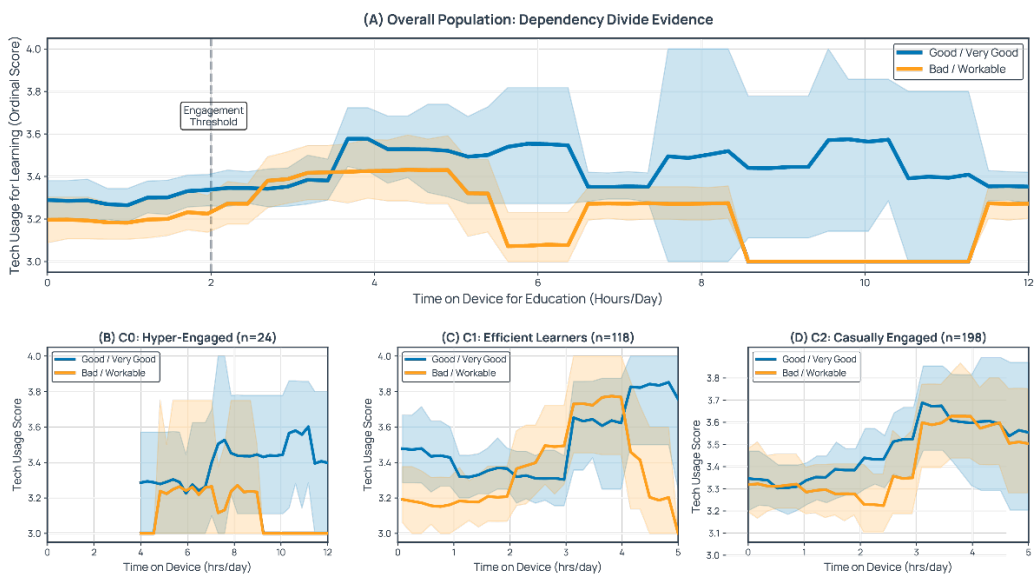

[Figure 3 – Interaction of TimeOnDeviceEducation and InternetReliability on TechForLearning, overall and by cluster.]

# Causal robustness checks: propensity score matching

Because students who spend more time on educational device use may differ systematically from their peers on demographic or socio-economic variables, the interaction could, in principle, be confounded by selection effects. To address this, the interaction between TimeOnDeviceEducation and InternetReliability for PlatformHelpfulness was re-estimated using propensity score matching (PSM), treating high versus low TimeOnDeviceEducation as the exposure. Propensity scores were estimated from Age, Gender, SES (FamilyExpenditure), ParentEducation, and Class, and 1:1 nearest-neighbor matching with a caliper of 0.024 (0.25 × SD of propensity scores) produced a matched sample of $n = 249$, corresponding to a 99.4% match rate.

Balance diagnostics (Table S1) show that all covariates achieved satisfactory post-matching balance, with standardized mean differences reduced and all post-matching p-values exceeding 0.05, indicating effective control of observed confounding. In this matched sample, the TimeOnDeviceEducation × InternetReliability interaction remained statistically significant for PlatformHelpfulness ($\beta \approx 0.208$, $p \approx 0.046$, 95% CI [0.004, 0.413]) and its magnitude modestly increased relative to the original OLS estimate (β ≈ 0.190), suggesting that the original result was, if anything, conservative. By contrast, the corresponding interaction for TechForLearning was not statistically significant in the matched sample (β ≈ 0.073, p ≈ 0.281), indicating that the dependency mechanism manifests more clearly in platform-specific satisfaction than in general technology-for-learning frequency.

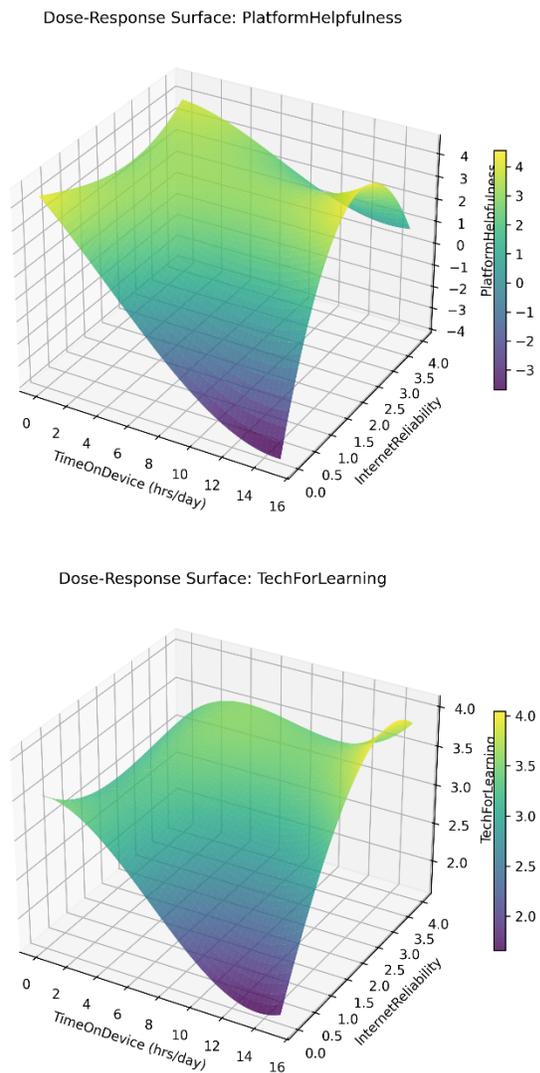

[Figure 4 – Forest plot comparing original and PSM-estimated interaction coefficients for PlatformHelpfulness and TechForLearning.]

Taken together, these PSM results strengthen the interpretation of the Dependency Divide as reflecting a plausible causal mechanism rather than a spurious association driven by demographic or socio-economic selection into intensive educational engagement. The pattern that the interaction persists and slightly increases in magnitude after matching, while the TechForLearning analogue attenuates, also clarifies the boundary conditions: the dependency effect is strongest for the platform-mediated aspects of the digital learning experience.

# Mechanistic pathway: mediation via perceived reliability

To probe the mechanism underlying the Dependency Divide, a mediation analysis was conducted with TimeOnDeviceEducation as the predictor, InternetReliability as the mediator, and PlatformHelpfulness as the outcome. In this model, higher educational device time was associated with slightly worse perceived reliability (Path A, β ≈ −0.059, p ≈ 0.022), and better perceived reliability in turn predicted higher platform helpfulness (Path B, β ≈ 0.169, p < 0.001). The product of these paths yielded a negative indirect effect of approximately −0.010 (95% bootstrap CI [−0.023, −0.002], p ≈ 0.002), corresponding to about 15% of the total effect of TimeOnDeviceEducation on PlatformHelpfulness.

This pattern suggests that as students become more engaged, they experience and notice infrastructure problems more acutely, which erodes their platform satisfaction even when the platform itself is otherwise functional. While the indirect effect is modest in magnitude, its statistical significance and theoretically coherent sign offer a concrete mediating pathway: deep engagement heightens sensitivity to reliability, and this heightened sensitivity partially transmits engagement's effect onto perceived platform usefulness. In combination with the interaction evidence, the mediation results depict dependency not simply as a high level of use but as a state in which engagement amplifies exposure to infrastructural fragility.

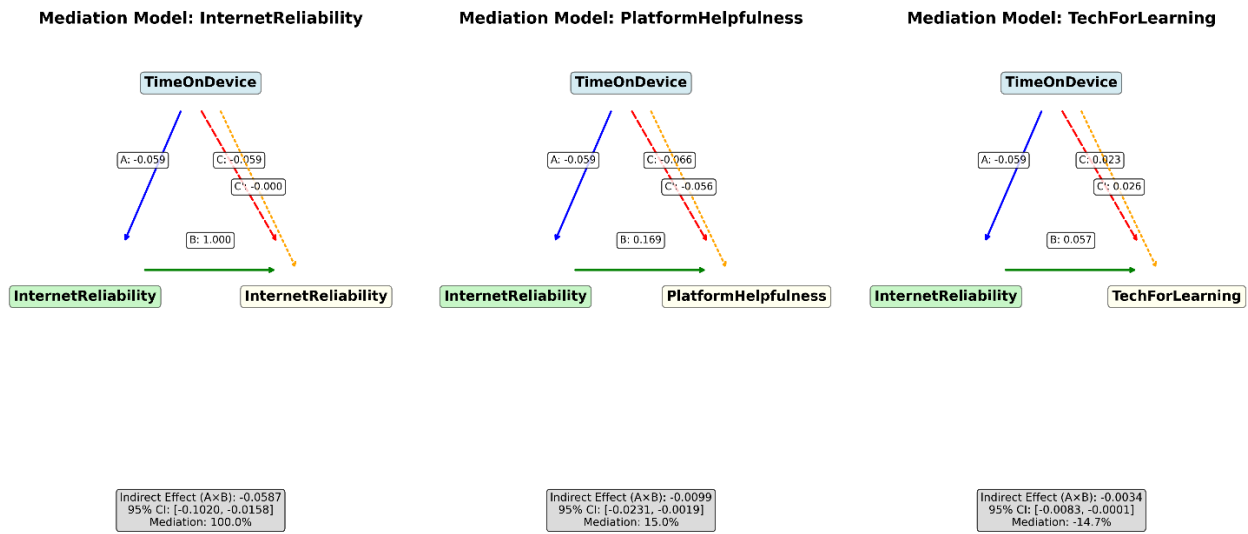

[Figure 5 – Mediation diagram with coefficients for Paths A and B and the indirect effect.]

# Policy-focused simulations and dose–response patterns (RQ3)

Finally, to connect the statistical findings to RQ3 on policy implications, counterfactual simulations were conducted using the best-performing supervised models to estimate the effect of improving InternetReliability by one standard deviation (≈0.95 points on the five-point scale) for different student profiles. For each cluster, predicted PlatformHelpfulness scores were computed under observed reliability and under a scenario where reliability was shifted upward for all students while holding other covariates fixed, and average treatment effects were then compared across profiles.

These simulations revealed pronounced heterogeneity in returns to infrastructure improvement: Efficient Learners exhibited the largest gains in PlatformHelpfulness (mean treatment effect ≈ 0.136, 95% CI [0.050, 0.221]), whereas Casually Engaged students showed more modest gains (≈ 0.020) and Hyper-Engaged students achieved intermediate absolute gains but the highest gains per unit of baseline satisfaction. Expressed as an efficiency ratio, infrastructure upgrades for Hyper-Engaged students yielded roughly 2.06 times the satisfaction improvement per student compared with a uniform intervention targeted at Casually Engaged students, underscoring the policy value of identifying dependency-prone subgroups. These heterogeneous treatment effects are visually apparent in a forest plot of cluster-specific effects and directly support targeted rather than one-size-fits-all infrastructure policies.

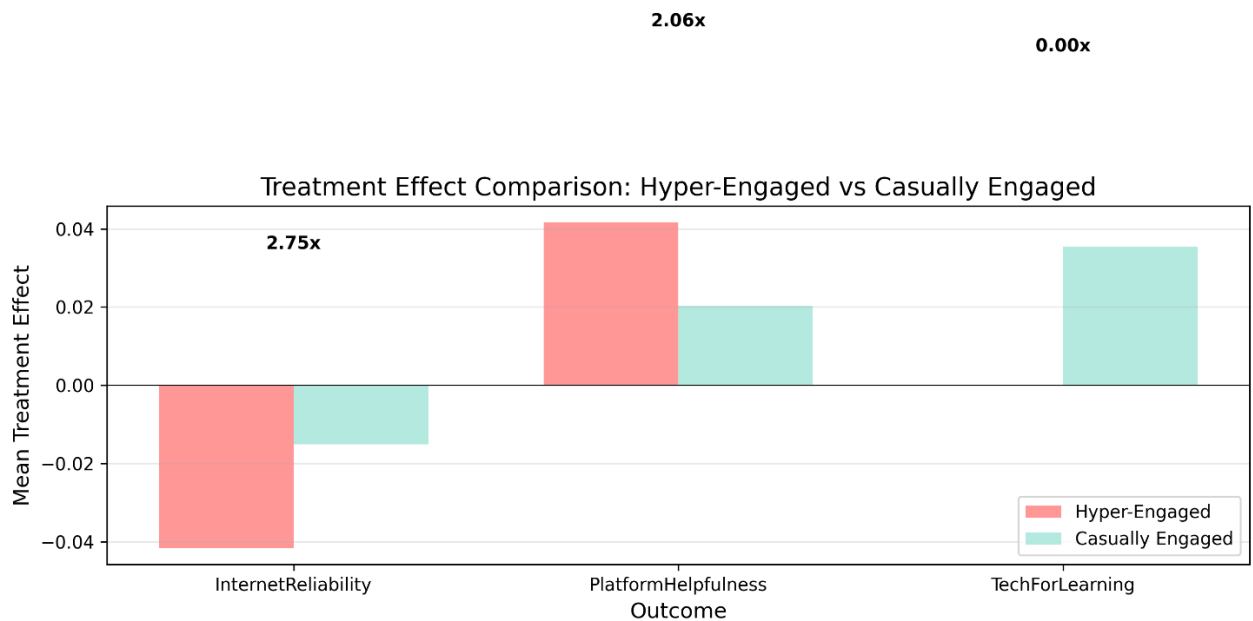

[Figure 6 – Cluster-specific treatment effects of a one SD improvement in InternetReliability on PlatformHelpfulness.]

In parallel, a dose–response analysis of educational device time under high versus low reliability conditions corroborated prior ALE-based thresholds, with PlatformHelpfulness plateauing at roughly 2.4 hours/day when reliability is good and showing little improvement or slight decline under poor reliability. This pattern suggests that beyond a moderate level of educational engagement, additional hours confer limited benefits unless accompanied by reliable infrastructure, reinforcing the argument that policy should focus simultaneously on raising constructive engagement among Casually Engaged students and securing stable connectivity for those already highly engaged. Together, the simulations and dose–response analyses translate the Dependency Divide from an abstract mechanism into concrete, cluster-specific levers for institutional decision-making in resource-constrained settings.

# Discussion

We developed and experimented with interpretable machine learning to study how access, use and use but out of pocket costs of quality of infrastructure interact and shape student digital satisfaction in a Global South university context. Using leakage-aware clustering, design-weighted supervised models, and targeted causal analyses, we motivate a new concept—Dependency Divide—derived explicitly out of the strong digital pressure points where the most digitally engaged students are rendered vulnerable in a fragile digital and infrastructural landscape. We extend first- and second-level digital divide accounts to show that satisfaction is not just a vector of access or usage intensity, but emergent upon the interaction of engagement and reliability.

# Measurement and construct implications

By psychometric standards, the four-item satisfaction scale is very low ($\alpha \approx 0.228$) when considered using ordinal-data methods, such that it is clear that the set of items are not combined into a single construct. Reliability estimates based on Spearman and polychoric both fail to exceed minimum thresholds, and diagnostics at the item level suggest that TechCollaboration is particularly divergent from the other items. The analytical conservatism of not amalgamating InternetReliability, PlatformHelpfulness, and TechForLearning into a composite score is warranted, although the clearest causal claim lies with our result that expresses "satisfaction with digital learning" as a multi-dimensional space consisting of infrastructure, platform experience, and behavioral use--each related to but distinct in capturing parts of students' digital lives.

# Interpretable segmentation: three student personas

The segmentation reveals three student personas that are statistically distinct, but also substantively distinct student bodies that should caution us around one-size-fits-all policy narratives about digital. Casually Engaged students are majority (51%): Students with moderate total screen time but the lowest educational use of those devices, and with access/attitude profiles that mix adequate access to sufficient quality of learning experiences (with ambivalence). Efficient Learners had a larger share of total time spent studying, were more satisfied with their digital learning experience, had a higher description of technology to utility fit within a task, which speaks to potential for better usability successes from improvements to platforms with these students. Hyper-Engaged students are a fringe population (11%): Extreme users of educational use of device (whom it is more difficult to classify with one label and are heterogeneous, although it is important to devote analytical attention toward), with complaints about infrastructure. Bootstrap stability indices argue for viewing the two larger clusters as core, with natural worry around the stability of Hyper-Engaged cluster arguing for the heterogeneity of that class. The qualitative themes from open-ended responses resonate with this persona construction, where Hyper-Engaged students repeat both the productivity gains related to dependency and also the emotional burden of dependency on unstable connectivity, whereas Casually Engaged students spoke more the quality of teaching and distractive nature of digital tools. Efficient Learners further emphasize the value of digital resources in clarifying difficult topics to study, consonant with their quantitatively measured moderate total time but efficient educational use relative to other students.

# Predictive modeling: modest accuracy, strong interpretability

The models, intentionally, do not perform excellently, with Quadratic Weighted Kappa values below 0.20 and Macro-F1 scores indicating class-wise discrimination hovering just above zero, suggesting that these are difficult data to predict at fine-grained ordinal outcomes on a relatively small, heterogeneous sample of respondents, while also a result of the analytical decision to privilege interpretability and design-weighting over aggressive tuning. Random Forest is the most consistent model for InternetReliability and PlatformHelpfulness; XGBoost performs best for TechForLearning, indicating that the different axes of satisfaction/toward which we are aiming the explanatory hits raise different predictive challenges. More importantly for the purposes of this paper, model-agnostic interpretation finds, consistently across outcomes, key drivers as TimeOnDeviceEducation, TimeOnDevice, Age, and access variables, with the former a persistent top predictor across outcomes and clusters of outcomes. ALE plots reveal non-linear dose-response type relations, in which educational device time can be beneficial up to about 2 times a day per day, after which the marginal benefit begins to decrease or turn negative depending upon outcome and reliability level. This modest predictive accuracy and stable interpretable driver plots fit an explanatory, policy-oriented agenda even if they take high-stakes individual prediction off of the table.

# The Dependency Divide: conditional vulnerability of engagement

The first theoretical contribution is to the Dependency Divide, the idea that engagement increases students' exposure to fragility/collapse associated with infrastructure, so that a highly engaged student is 'at risk' for collapse and its consequences when reliability is poor. Interaction modeling shows that the product (interaction) of education device time and internet reliability together subsume and are more informative than the main effects of each: the time on device toward education × internet reliability interaction is positive and statistically significant, but both main effects are non-significant, of the TechForLearning 'experience' model. This shows that neither "more engagement" nor "better infrastructure" is sufficient in isolation—satisfaction only increases when a lot of engagement meets with adequate reliability. Predicted margins illustrate this visually: with connectivity is good or very good, the more time students spend exposed to educational devices, the higher their TechForLearning total score, but only below a certain cut-off (up to a moderate number of hours, rather than an infinite straight line), and with bad or merely workable connectivity, the more hours they have, the less value they obtain, and they're likely to feel more frustrated than on a straight line. Cluster-stratified plots reveal that the total gap between high and low trajectories for reliability is steepest for the most Hyper-Engaged—but all measure true". Collectively, these results suggest not just the concept of the Dependency Divide, but more, i.e., a sub-quantification of it based on dual behavioral and experiential data.

# Strengthening causal interpretation

The causal analyses provide evidence that the Dependency Divide can be interpreted as a causal mechanism rather than mere correlation. Specifically, propensity score matching addresses the concern that students who invest heavily in educational device use might differ systematically from their peers in ways that jointly affect both engagement and satisfaction. After matching students on age, gender, socio-economic status, parental education, and class using 1:1 nearest-neighbor matching with a tight caliper, covariate balance is attained across all observed confounders. In this matched sample, the TimeOnDeviceEducation × InternetReliability interaction for PlatformHelpfulness remains statistically significant and is even slightly larger in magnitude relative to the OLS estimate from the unmatched sample, suggesting that the original estimate was if anything conservative.

Perhaps more informative, the matched-sample analysis reveals a boundary condition: the interaction for TechForLearning is not statistically significant after balancing covariates, suggesting that the Dependency Divide is more pronounced for platform-specific satisfaction than for general technology-for-learning frequency. The specificity of this finding strengthens the theoretical account, since it localizes the dependency mechanism at the level where infrastructure and platform design directly intersect—students' lived experience of whether the institutional systems they must rely on actually work when they need them to. In sum, we take the PSM results as supporting a tentative causal interpretation: when students are already engaged, increasing reliability causes higher platform satisfaction, whereas low reliability diminishes the returns to engagement.

## Mechanism: mediation via perceived reliability

This mediation analysis provides a bit more detail about one of the ways in which we think our causal story may work. As time on educational device increases students' scores on InternetReliability get worse (Path A): more time spent using educational devices exposes them to more or more obvious signs of infrastructure failing. More reliable perceptions in turn predict much higher scores in PlatformHelpfulness (Path B), with a statistically significant indirect effect that accounts for 15% of the total effect of engagement on platform satisfaction. Although small, the direction and significance of the indirect effect fits both the interaction analysis, and accounts from two different Hyper-Engaged students of being "trapped" by unreliable things that they still need to use a lot. Our mediation model suggests that this dependency is not only one of relying on fragile infrastructure—sometimes described in the Hyper-Engaged students' accounts—but also one of experiencing it: students become more aware of infrastructure problems the more they engage, and part of the effect that engagement has on platform value is passed along the transmission path of having more acute awareness of infrastructure.

## Policy implications: targeting reliability where dependency is highest

The policy simulation and dose-response analysis take those statistical patterns and turn them into institutional decision-making actions. Simulating counterfactuals (what would have happened) from a one-standard-deviation increase in InternetReliability shows that Efficient Learners have the greatest positive impact on PlatformHelpfulness as compared to other groups; and, that Hyper-Engaged students have the greatest rate of return on investment per student compared to Casually Engaged students. If we express those returns as an efficiency ratio, then investing in reliability to benefit Hyper-Engaged students would be more than two times as cost-effective to raise satisfaction as implementing similar reliability enhancements to benefit all Casually Engaged students. In addition, the dose-response analysis shows that satisfaction on the platform reaches its maximum point (plateau) at approximately 2.4 hours of daily educational device usage, when reliability is good; and at much lower levels of usage when reliability is poor. Overall, these results support a precision policy strategy: institutions should concurrently promote moderate engagement among Casually Engaged students, allow

Efficient Learners to continue to use digital platforms with reliability to focus their digital practices, and invest in reliable technology for the strategic group of Hyper-Engaged students who are most dependent on digital systems. Instead of defining expansion of access as an end in itself, this perspective suggests that institutions should define reliability of infrastructure in relation to the level of dependency of each student.

# Limitations and directions for future work

Several limitations qualify these conclusions and point to directions for future research. First, our study is based on cross-sectional self-report data from a single public university, which constrains both causal inference and external generalizability (despite leveraging design weights and PSM). Longitudinal designs that track how engagement, infrastructure, and other outcomes change over time would offer stronger causal identification and study of the trajectories of dynamic dependencies. Second, sample size—especially for the particularly highly engaged cluster—limits the precision of cluster-specific estimates, and contributes to the modest predictive performance of the supervised models. Larger samples, perhaps across multiple institutions or regions, would permit finer grained profiles and more ambitious modeling without losing interpretability. Third, while we employ mixed-methods integration and interpretable ML, we are still left relying on a finite set of survey items that certainly do not capture the full complexity of students' digital ecologies (for example, device-sharing arrangements, platform-specific features, or psychosocial factors like digital fatigue). Future studies might triangulate the mechanisms behind the Dependency Divide more fully via combinations of log data, richer qualitative work, and experimental manipulation of platform features or support services. Finally, in the face of exactly such a danger, as critical learning analytics scholars have urged, profiling any individual student carries some risk of reinforcing inequities if done uncritically in policy or practice. The diagnostic framework we have advanced here is understood to apply principally or best to helping identify where reliability failures do the most harm: not as grounds for restricting access or shifting responsibility for infrastructural failures onto students. Overall, our work points to the ways in which in a post-access world, the key question is less about "who has technology" than it is about "who must depend on fragile infrastructure for their learning, and what happens when that infrastructure falls short." By rigorously operationalizing and testing what we call the Dependency Divide, we hope to shed light on both a conceptual lens and a method "toolkit" that institutions might use to help align their digital investments with the reality of their most vulnerable—and most engaged—students.

# Conclusion

This study shows that when using educational technology in resource-limited environments, an individual can be conditionally vulnerable based on the amount of effort used to engage with the technology. A new concept called the Dependency Divide has been developed as a result of this analysis. The study was conducted with 396 Bangladeshi university students who were categorized into three different types of learner types. Hyper-Engaged learners are defined as learners that both use the most technology and spend the most time studying; however, for these students, only when the internet is functioning at a high level does a positive relationship exist between the amount of time spent studying and technology use (β = .033, p = .028) as compared to the relationship between study time and technology use for Casual Learners and Efficient Learners. These results show that current assumptions made in technology acceptance studies regarding engagement as being uniformly advantageous may be incorrect.

Three distinct and relatively stable student profiles were identified as a result of the cluster analysis—Casually Engaged (58%), Efficient Learners (35%), and Hyper-Engaged (7%). In addition to providing a descriptive categorization of learner types, the existence of three distinct profiles allows for an examination of the differences in vulnerability as a function of learner type. Specifically, Casually Engaged students are satisfied with their learning experiences regardless of the level of quality of the infrastructure supporting those experiences, primarily due to the fact that the workflow used by these students is less dependent on continuous connectivity. Conversely, Hyper-Engaged students experience feelings of "helplessness" when their connection fails during live sessions. Capability, therefore, can become liability. This reversal of capability having the potential to become liability is important, as it indicates that well-meaning digital transformation initiatives may actually exacerbate inequalities by creating dependency relationships that are unmanageable for fragile infrastructures to support.

From a methodological standpoint, the leakage-aware framework utilized in this study addresses a long-standing issue in educational data mining, specifically the issue that the variables measured as outcomes often influence the clustering process. As such, the satisfaction measures were excluded from the clustering input data and the stability of the clusters was validated through 1000 iterations of bootstrapping. Therefore, the resulting profiles represent actual usage patterns rather than levels of satisfaction. In addition, the utilization of K-prototypes clustering in conjunction with profile-specific random forests and SHAP interpretation provide a replicable model for researchers wishing to extract actionable information from mixed-type educational data. Furthermore, this approach emphasizes the importance of explanatory power as opposed to predictive accuracy in order to allow institutional leaders to understand why their intervention efforts are successful and for whom, as opposed to determining whether their predictive models accurately forecast future events.

Policy-wise, there are three possible strategic actions that can be taken as a result of this research. First, infrastructure investment decisions should focus on ensuring that high dependency users have access to reliable and consistent connections as opposed to investing in increased access to infrastructure for all users. This is a feasible decision-making strategy given limited budgets. Second, universities should develop emergency response plans that enable Hyper-Engaged students to continue their learning activities uninterrupted despite outages (i.e., offline content availability, alternative delivery methods, etc.). Finally, student awareness campaigns should be designed to educate students on recognizing and managing dependency risk and adapt their workflows to varying levels of connectivity as opposed to relying on the assumption of reliable infrastructure.

In addition to the above-mentioned findings, there are also some limitations in this research. The number of participants in this research is limited to only one university from Bangladesh, thus it could not be generalized to all educational institutions and/or national context. The cross-sectional design of the study limits the researchers' ability to make causal statements regarding the development of dependency over time. For example; do students develop into Hyper-Engaged as a result of their reliable access to the Internet or are Hyper-Engaged students frustrated by reliability of connectivity to use less technology? To determine this longitudinal data would be needed. Furthermore, although the statistical significance of the interaction effects was found ($\beta = .033$) the magnitude of the interaction effect was very small, suggesting that there are many other variables that influence the users level of satisfaction in relation to their level of engagement with an application, and that the quality of the infrastructure is simply one variable that has a theoretical impact on the relationship between engagement and satisfaction. Additional areas of research include investigating whether the Dependency Divide exists in other resource-constrained contexts and domains beyond education. For example, does high engagement with telemedicine platforms create a similar level of vulnerability in regions with unreliable connectivity? The theoretical mechanism underlying the Dependency Divide—the workflows optimized for reliability will fail catastrophically under conditions of instability—may operate in a variety of contexts where digital transformations occur in fragile infrastructure environments. Investigating how users respond to varying levels of reliability in terms of either developing resilience strategies or reducing dependence on the technology also represents a critical area of investigation. The Dependency Divide framework serves as a lens for understanding not whether digital access benefits users, but when and for whom the benefits of digital access materialize.

# Funding Statement

This research received no specific grant from any funding agency in the public, commercial, or not-for-profit sectors.

# Conflict of Interest Statement

The authors declare no competing interests.

# Author Contributions

MMF: Conceptualization, Data collection, Methodology, Formal analysis, Writing - original draft

HA: Data collection, Investigation, Writing - review & editing

MMH: Supervision, Validation, Writing - review & editing

MRK: Conceptualization, Supervision, Writing - review & editing

# Acknowledgments

We thank the students who participated in this study and the university administration for facilitating data collection. We also thank anyone who provided feedback/assistance.